\documentclass[letterpaper, 10 pt, conference]{ieeeconf}  

\IEEEoverridecommandlockouts                              

\usepackage[dvipsnames, table]{xcolor}
\usepackage{graphicx}
\usepackage{siunitx}
\DeclareSIUnit{\nothing}{\relax}
\usepackage{booktabs}
\usepackage{hyperref}
\usepackage{multirow}
\usepackage{xcolor}
\usepackage{comment}

\hypersetup{%
    pdfborder = {0 0 0}
}
\DeclareSIUnit{\nothing}{\relax}
\usepackage[super]{nth} 

\title{\LARGE \bf
A Deep Learning-based Pest Insect Monitoring System for Ultra-low Power Pocket-sized Drones
}

\author{Luca Crupi$^{1}$, Luca Butera$^{1}$, Alberto Ferrante$^{1}$, and Daniele Palossi$^{12}$
\thanks{$^{1}$Luca Crupi, Luca Butera, Alberto Ferrante, and Daniele Palossi are with the Dalle Molle Institute for Artificial Intelligence (IDSIA), USI and SUPSI, Lugano, 6962, Switzerland. 
        {\tt\small name.surname@usi.ch}}%
\thanks{$^{2}$Daniele Palossi is also with the Integrated Systems Laboratory (IIS), ETH Z\"urich, Z\"urich, 8092, Switzerland.
        }%
}

\begin{document}
\maketitle

\begin{abstract}
Smart farming and precision agriculture represent game-changer technologies for efficient and sustainable agribusiness.
Miniaturized palm-sized drones can act as flexible smart sensors inspecting crops, looking for early signs of potential pest outbreaking.
However, achieving such an ambitious goal requires hardware-software codesign to develop accurate deep learning (DL) detection models while keeping memory and computational needs under an ultra-tight budget, i.e., a few \SI{}{\mega\byte} on-chip memory and a few 100s \SI{}{\milli\watt} power envelope.
This work presents a novel vertically integrated solution featuring two ultra-low power System-on-Chips (SoCs), i.e., the dual-core STM32H74 and a multi-core GWT GAP9, running two State-of-the-Art DL models for detecting the Popillia japonica bug.
We fine-tune both models for our image-based detection task, quantize them in 8-bit integers, and deploy them on the two SoCs.
On the STM32H74, we deploy a FOMO-MobileNetV2 model, achieving a mean average precision (mAP) of 0.66 and running at \SI{16.1}{frame/\second} within \SI{498}{\milli\watt}.
While on the GAP9 SoC, we deploy a more complex SSDLite-MobileNetV3, which scores an mAP of 0.79 and peaks at \SI{6.8}{frame/\second} within \SI{33}{\milli\watt}.
Compared to a top-notch RetinaNet-ResNet101-FPN full-precision baseline, which requires 14.9$\times$ more memory and 300$\times$ more operations per inference, our best model drops only 15\% in mAP, paving the way toward autonomous palm-sized drones capable of lightweight and precise pest detection.
\end{abstract}


\section{Introduction} \label{sec:intro}

A crucial aspect of any modern farming activity is a precise and timely intervention in the case of pest (insect) infestations to minimize the production/economic damage and the environmental impact of the required treatments~\cite{DLEntomology}.
For example, early identification of harmful bugs can lead to ad-hoc treatments, such as spraying only part of the cultivated field or a few plants, up to highly accurate treatments of only part of single trees/plants.
Conversely, traditional mass-scale farming productions have adopted coarse-grained strategies, spraying chemicals in the entire cultivated field, even in isolated or partial infestations.
An important step forward in the accuracy of treatments has been possible by disseminating traps in the cultivated area to monitor the presence of specific pests, such as Cydia pomonella~\cite{JU2021104925} or Popillia japonica~\cite{EPPO}.
However, early approaches required costly and time-consuming expert human operators' intervention to manually inspect the traps and collect information about the condition of the crop~\cite{preti2021insect}.

In the last decade, thanks to the advent of Internet-of-Things (IoT) technologies, the diffusion of embedded ultra-low power Systems-on-Chips (SoCs) and sensors' miniaturization, trap-based precision agriculture has become smarter by automating the monitoring procedure with embedded devices integrated on the traps~\cite{SmartTrap}.
These battery-powered systems are usually composed of an image sensor (e.g., color or infrared camera), a low-power microcontroller unit (MCU), and a radio (e.g., WiFi, GSM, 4/5G) to stream data (e.g., images) to a power-unconstrained remote server for the analysis.
Despite the paramount improvement compared to human-operated traps, these State-of-the-Art (SoA) solutions still require costly external infrastructure, i.e., mainframes and servers, and need high-throughput radio connectivity with the additional disadvantage of draining small-capacity batteries typically available on the traps~\cite{SmartTrap}.

\begin{figure}[tb]
\centering
\includegraphics[width=1.0\columnwidth]{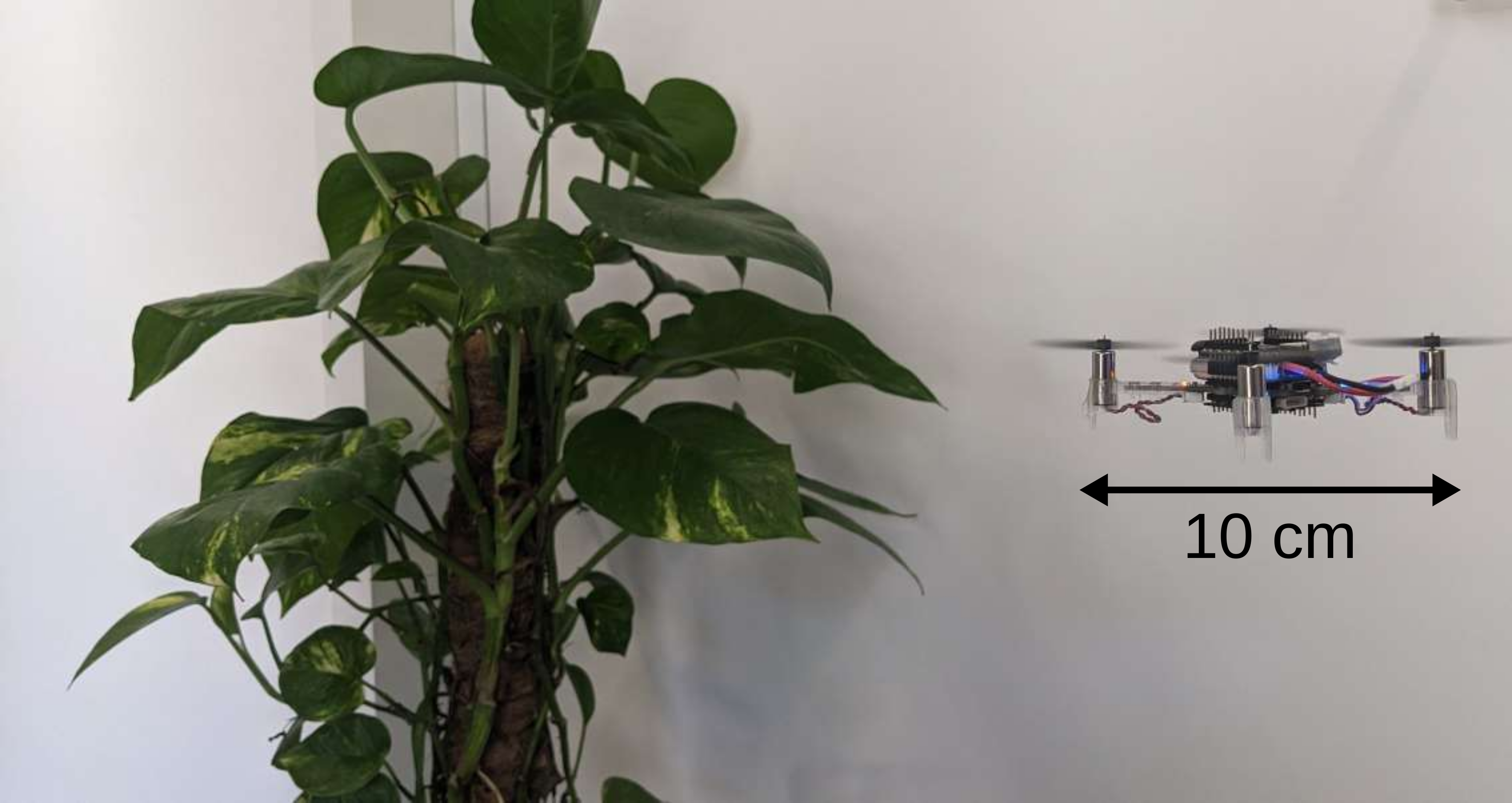}
\caption{Use case: a pocket-sized nano-drone inspecting a plant relying only on onboard sensing and computational capabilities. }
\label{fig:nano-drone}
\end{figure}

\begin{table*}[t]
    \small
    \caption{Network survey (number of parameters, operations, throughput, and mAP) on the models used in~\cite{9601235} and extended with two additional ones introduced in our work (\textbf{in bold}). The GPU employed is an Nvidia GeForce RTX 2080.}
    \label{tab:other_models_params}
    \resizebox{\linewidth}{!}{%
    \begin{tabular}{llcccccc}
    \toprule
    \textbf{Network} & \textbf{Input size} & \textbf{\# Param. [\SI{}{\mega\nothing}]} & \textbf{\# Op. [\SI{}{\giga MAC}]} & \textbf{Device} & \textbf{Frame-rate [\SI{}{\hertz}]} & \textbf{mAP}\\
    \toprule
    FasterRCNN-VGG16-FPN & $800\times800\times3$ & 31.90 & 275.23 & GPU & 11.90 & 0.92\\
    FasterRCNN-ResNet101-FPN & $800\times800\times3$ & 60.20 & 167.41 & GPU & 11.23 & 0.92\\
    FasterRCNN-DenseNet169-FPN & $800\times800\times3$ & 30.00 & 73.99 & GPU & 7.59 & 0.91\\
    FasterRCNN-MobileNetV3-FPN & $800\times800\times3$ & 18.90 & 18.41 & GPU & 60.92 & 0.93\\
    \midrule
    RetinaNet-VGG16-FPN & $800\times800\times3$ & 22.90 & 270.06 & GPU & 12.37 & 0.91\\
    RetinaNet-ResNet101-FPN&$800\times800\times3$&51.20&174.85 &GPU&11.83&0.93\\
    RetinaNet-DenseNet169-FPN&$800\times800\times3$&21.00&65.10 &GPU&18.49&0.92\\
    RetinaNet-MobileNetV3-FPN&$800\times800\times3$&10.60&11.09 &GPU&48.91&0.91\\
    \midrule
    SSD-VGG16&$300\times300\times3$&12.10&26.58 &GPU&42.58&0.56\\
    SSD-ResNet101&$300\times300\times3$&32.30&42.39 &GPU&29.76&0.87\\
    SSD-DenseNet169&$300\times300\times3$&11.80&24.51&GPU&28.59&0.92\\
    SSD-MobileNetV3&$300\times300\times3$&7.50&3.42&GPU&33.20&0.80\\
    \midrule
    \textbf{SSDLite-MobileNetV3}&\textbf{320$\times$240$\times$3}&\textbf{3.44}&\textbf{0.58}&\textbf{GAP9}&\textbf{6.8}&\textbf{0.80}\\
    \textbf{FOMO-MobileNetV2}&\textbf{96$\times$96$\times$3}&\textbf{0.02}&\textbf{0.01}&\textbf{STM32}&\textbf{16.1 }&\textbf{0.66}\\
    \bottomrule
    \end{tabular}
    }
\end{table*}

In this context, our work provides a vertically integrated system for accurate pest detection employing ultra-constrained embedded MCUs, i.e., within a few 100s \SI{}{\milli\watt} power envelope and a few \SI{}{\mega\byte} on-chip memory, and cheap low-resolution cameras.
Leveraging SoA deep learning models for detecting Popillia japonica bugs, we present a novel hardware-software co-design that is an ideal fit for both traditional traps and aboard miniaturized palm-sized nano-drones, which can autonomously and dynamically inspect crops, as drafted in Figure~\ref{fig:nano-drone}. 
Employing nano-drones, such as miniaturized blimps~\cite{palossi2017self} in greenhouses or autonomous quadrotors~\cite{10342162}, for this type of monitoring and detection activities has the additional advantage of flexibility. 
For example, these tiny (less than \SI{10}{\centi\meter} in diameter) robotic platforms can easily reach locations where traditional traps are not deployable or unreachable by bulky robotic arms, e.g., attached to trucks and tractors.

Therefore, to achieve our goal, we must face both the challenges posed by hardware-constrained devices and develop robust deep-learning models capable of distinguishing between small dangerous bugs and harmless species (sometimes only a few pixels in blurry images).
For example, the Popillia japonica we address in this work has similar visual features to the innocuous Cetonia aurata and Phyllopertha horticola, shown in Figure~\ref{fig:datasetSamples}.

On the one hand, we select two embedded devices, both compatible with the tiny power envelope of battery-powered traps and nano-drones: the Arduino Portenta H7 (STM32H74 dual-core MCU) and the Greenwaves Technologies (GWT) multi-core GAP9 SoC.
On the other hand, we survey the field of deep learning models for pest detection, and we select two SoA models, a powerful SSDLite-MobileNetV3~\cite{howard2019searching} and a lighter FOMO-MobileNetV2 model~\cite{sandler2018mobilenetv2}.
The former is a top-scoring convolutional neural network (CNN), which can run on the GAP9 SoC, while the latter is more suitable for the STM32H74 MCU.
In detail, our contributions can be summarized as follows:
\begin{itemize}
    \item We start from models pre-trained on the COCO dataset for object detection~\cite{cocodataset}, and we then perform an additional fine-tuning stage with a custom dataset (\SI{3}{\kilo\nothing} images) for the Popillia japonica detection.
    \item We quantize, implement, and deploy both models on the two SoCs. On the STM32H74, we execute and profile the FOMO-MobileNetV2 in float32 full-precision and its quantized 8-bit version. On the GAP9, instead, we deploy the SSDLite-MobileNetV3 in float16 and int8 (quantized), comparing the performance between the general-purpose multi-core domain of the GAP9 and its convolutional hardware accelerator (called NE16).
    \item Finally, we perform a thorough assessment of both classification performances and on-device inference, including power measurements on the two target boards.
\end{itemize}

Our results highlight how both CNNs can achieve a top-notch mean average precision (mAP) of 0.66 for the FOMO-MobileNetV2 and 0.79 for the SSDLite-MobileNetV3, with minimal loss w.r.t. SoA solutions~\cite{9601235} (i.e., minimum loss of 0.01 of mAP) even in the case of strong 8-bit quantization.
Furthermore, the FOMO-MobileNetV2 running on the Portenta peaks with a throughput of \SI{17.5}{frame/\second}@\SI{494}{\milli\watt}, while the SSDLite-MobileNetV3 deployed on the GAP9 board achieves \SI{4}{frame/\second}@\SI{31.4}{\milli\watt} and \SI{6.8}{frame/\second}@\SI{33}{\milli\watt}, running on the multi-core cluster and the NE16, respectively.
Therefore, the proposed systems represent a viable option for long-lasting smart traps (up to 283 days with a \SI{1000}{\milli\ampere\hour}@\SI{3.7}{\volt} battery) and aboard nano-drones.
\section{Related work} \label{sec:related_work}

Insect detection systems, using images and deep learning models, mostly leverage SoA object detectors, as surveyed in~\cite{PestDetectionSurvey}. 
For example, \cite{FasterRCNNInsect2} leverages a FasterRCNN-based detector~\cite{FasterRCNN}, while \cite{SSDInsect1} uses a Single Shot Multibox Detector (SSD)~\cite{SSD}, \cite{RetinaNetInsect1} uses RetinaNet~\cite{RetinaNet}, and \cite{RFCNInsect1} employs R-FCN~\cite{RFCN}, for detecting small insects.
Table~\ref{tab:other_models_params} reports the models studied in~\cite{9601235} extended by two smaller MobileNet-based models~\cite{sandler2018mobilenetv2, howard2019searching} employed in our work (last two \textbf{bold} lines).
This broad comparison considers the size of the input image, the number of parameters, the computational cost in MAC operations, the device used for testing the model, and the detection mAP with the same testing dataset used in our work.
Even though most networks perform remarkably well in mAP, i.e., up to 0.933, they require desktop-class devices to achieve a real-time throughput (max \SI{61}{frame/\second}).
This power-hungry class of computational devices is clearly unsuitable for IoT battery-powered smart traps or nano-drones due to their power requirement ($\sim$\SI{100}{\watt}) and weight ($\sim$\SI{1}{\kilo\gram}). 

As for other IoT applications, images are often collected using camera-equipped IoT nodes, but the actual pest detection algorithms run in the cloud~\cite{agriculture12101745}. 
Nodes deployed in the fields need to be cheap - and, thus, extremely limited in computational resources and memory - and have limited energy requirements. 
These nodes are often powered by batteries that can be recharged through small solar panels.
This solution greatly limits the monitoring frequency of the insects, as it would imply the frequent transmission of relatively large images (e.g., 640$\times$480 pixels images~\cite{SmartTrap}) over limited-bandwidth networks, with its associated energy usage. 
For example, in the iSCOUT\footnote{https://metos.global/en/iscout/} insect monitoring system, insect count can run, at most, three times a day. 
This choice of processing images in the cloud is often motivated by reasons other than technical (e.g., the collection of images for improving the models), even in applications where edge devices would have sufficient computational resources and energy for running the models.

Instead, considering autonomous detection systems without any need for remote computation and power-hungry communication streams, recent works allowed the deployment of object detection algorithms on the edge~\cite{humes2023squeezed, rusci2023parallel, 10137154}.
Squeezed edge YOLO~\cite{humes2023squeezed} is a network that performs object detection within less than 1 million parameters and is deployable on very limited parallel ultra-low power SoC.
\cite{rusci2023parallel} introduces a Viola-Jones-based algorithm for detecting the Cydia pomonella insect, running a battery-powered embedded system mounted on conventional traps.
Despite the simplicity of this approach, using pre-trained patches on a reduced amount of data to detect insects, it peaks at \SI{2.5}{frame/\second} a GWT GAP8 SoC. 
Moreover, the approach is not suitable for solving our task since we do not want only to detect but also to classify and discriminate different insects.
\cite{BETTISORBELLI2023108228} explores a CNN-based classification system that relies on a YOLO network to classify insects on drone images. 
Although accurate (up to 0.92 of mAP), the method struggles to classify clusters of insects correctly due to the suppression method that excludes a priori detections with high intersection over union.

Finally, \cite{10137154} presents an SSD-MobileNetV2 object detection algorithm fully deployed on the GAP8 SoC running aboard an autonomous nano-drone to detect tin cans and bottles.
This system reaches up to \SI{1.6}{frame/\second} with an mAP of 0.5.
Similarly, in~\cite{9401730}, the authors employ the GAP8 SoC for an automatic license plate recognition system using a MoileNetV2 with an SSDLite detector.
Similarly, also our CNNs are based on MobineNets~\cite{howard2017mobilenets}, and for the one we deploy on the GAP9 SoC, we adopt an SSDLite head achieving a peak throughput of \SI{6.8}{frame/\second}.
Ultimately, our work allows precise insect detection and classification tasks on computationally constrained devices that can be used in smart battery-powered IoT traps or aboard autonomous nano-drones.
\section{System implementation} \label{sec:implementation}

In this section, we discuss our two system designs featuring two ultra-low power embedded devices, i.e., a widely used dual-core Arduino Portenta H7 and a novel multi-core GWT GAP9 SoC featuring the NE16 convolutional hardware accelerator.
Then, given these two devices' significant memory and computing power differences, we explore two alternative SoA CNNs for pest detection: a lightweight FOMO-MobileNetV2~\cite{sandler2018mobilenetv2}, which we deploy on the Portenta, and a more complex SSDLite-MobileNetV3~\cite{sandler2018mobilenetv2, howard2019searching} for the more capable GAP9 SoC -- almost 100$\times$ more MAC operations per inference than the FOMO-MobileNetV2.

\subsection{Platforms} \label{subsec:platforms}

\begin{figure}[tb]
\centering
\includegraphics[width=1.0\columnwidth]{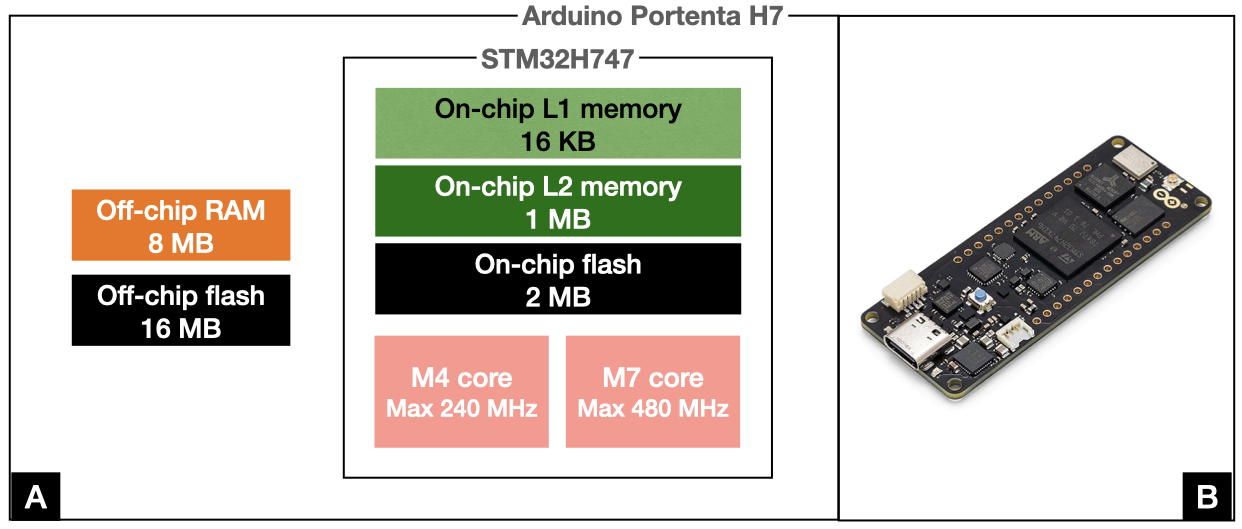}
\caption{Arduino Portenta H7 block diagram (A) and picture of the board (B).}
\label{fig:portenta}
\end{figure}

\begin{figure*}[tb]
\centering
\includegraphics[width=1.0\linewidth]{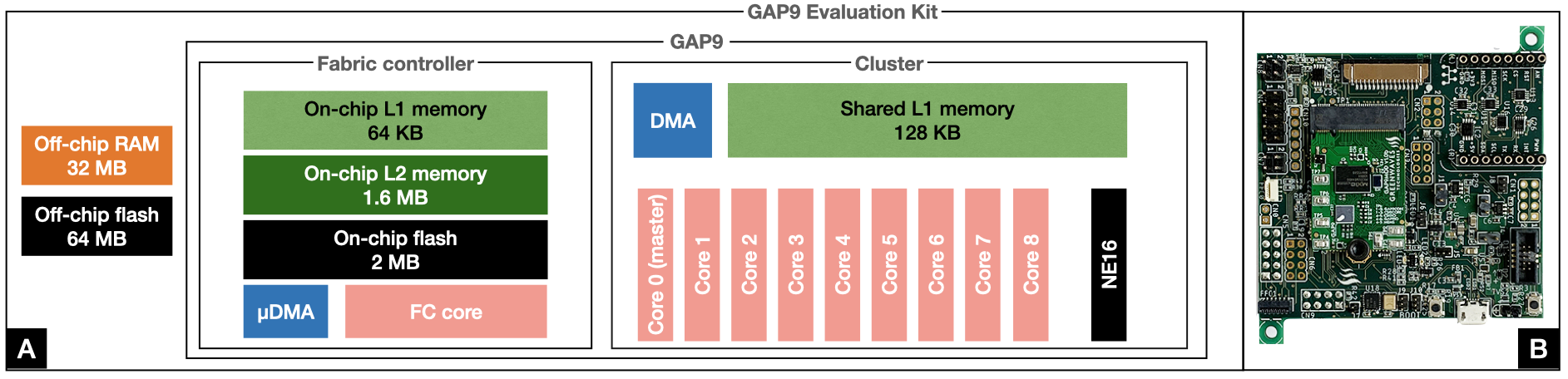}
\caption{GAP9 evaluation kit block diagram (A) and picture of the board (B).}
\label{fig:gap9evkit}
\end{figure*}

The detection systems we propose are based on two widely different platforms: the Arduino Portenta H7 board and the GAP9 evaluation kit.
The Arduino Portenta board is extended with the Portenta vision shield, which provides a grayscale, ultra-low power Himax HM-01B0 camera with a maximum resolution of 320$\times$320 pixels, which is also available on the GAP9 development kit.
The main differences between the two boards arise when comparing their computational resources. 
The Arduino Portenta, depicted in Figure~\ref{fig:portenta}, features an STM32H747 SoC with two cores: a Cortex\textregistered M4, running at a \SI{240}{\mega\hertz}, and a Cortex\textregistered M7, running at \SI{480}{\mega\hertz}.
The two cores can communicate via a \textit{remote procedure call} mechanism.
Both cores have a double-precision floating point unit (FPU) that allows the computations to be performed directly in floating point arithmetic.
The Portenta board features two off-chip memories, a \SI{16}{\mega\byte} FLASH and a \SI{8}{\mega\byte} RAM.
The STM32H747's memory hierarchy instead is composed by a \SI{2}{\mega\byte} FLASH, a \SI{1}{\mega\byte} L2 memory, and a \SI{16}{\kilo\byte} L1 memory.

\begin{table}[!h]
    \small\centering
    \caption{Comparison between the STM32H747 and the GAP9.\label{tab:comparisonArchitectures}}
    \addtolength{\tabcolsep}{10pt}
    \resizebox{\columnwidth}{!}{%
    \begin{tabular}{lll}
    \toprule
    \textbf{}&\textbf{STM32H747}&\textbf{GAP9}\\
    \midrule
    \textbf{General-purpose cores}&2&10\\
    \textbf{Frequencies [\SI{}{\mega\hertz}]}&480&370\\
    \textbf{CNN accelerators}&None&NE16\\
    \textbf{FPU availability}&Yes&Yes\\
    \textbf{On-chip L1 memory [\SI{}{\kilo\byte}]}&16&128\\
    \textbf{On-chip L2 memory [\SI{}{\mega\byte}]}&1&1.6\\    
    \textbf{On-chip FLASH [\SI{}{\mega\byte}]}&2&2\\
    \textbf{Off-chip RAM [\SI{}{\mega\byte}]}&8&32\\
    \textbf{Off-chip FLASH [\SI{}{\mega\byte}]}&16&64\\
    \bottomrule
    \end{tabular}
    }
\end{table}

The GAP9 evaluation kit, depicted in Figure~\ref{fig:gap9evkit}, features a GAP9 ultra-low power SoC that relies on ten RISC-V cores equipped with single-precision FPUs.
This SoC is characterized by a power domain featuring one core, called \textit{fabric controller}, which orchestrates the work for the multi-core cluster and acts as the interface with the external peripherals.
The cluster, instead, is meant for general-purpose intense parallel workloads, such as CNNs.
This powerful processor can deliver up to \SI{15.6}{\giga Op/\second} and has an on-chip L2 memory of \SI{1.6}{\mega\byte}.
The off-chip memories available on the GAP9 development kit (board) are \SI{32}{\mega\byte} RAM and \SI{64}{\mega\byte} FLASH.
Furthermore, the GAP9 SoC features the NE16, an accelerator specifically tailored for linear algebra computations that can consequently accelerate CNN operations but can be used only with int8 arithmetic.
It can reach up to 150 8-bit MAC operations per second.
Table~\ref{tab:comparisonArchitectures} summarizes the two boards' specifications.

\subsection{Neural networks}

\begin{figure}[tb]
\centering
\includegraphics[width=1.0\linewidth]{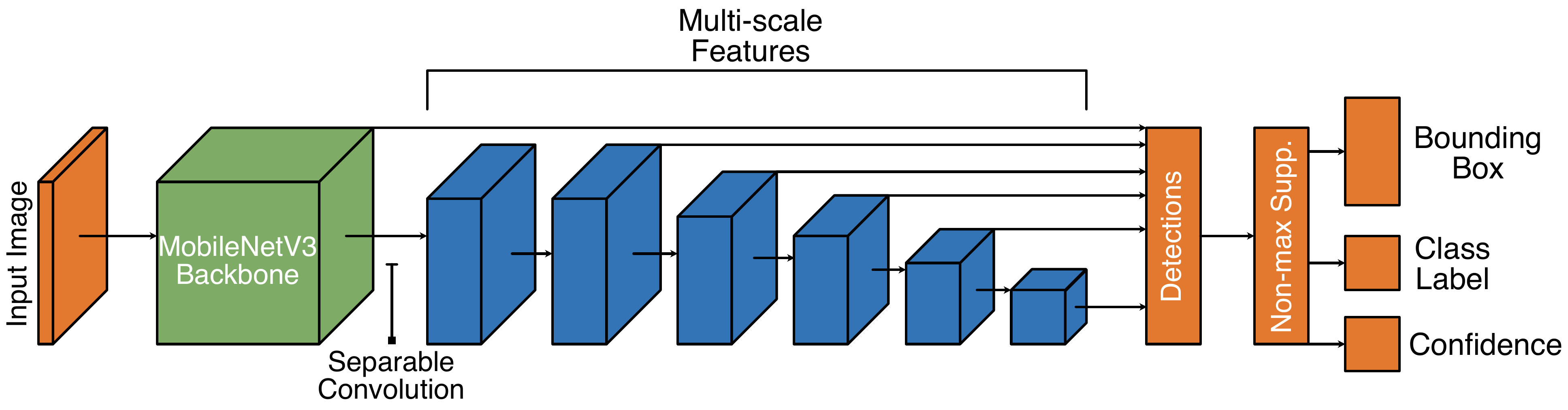}
\caption{\label{fig:ssdlite_mnet}SSDLite with MobileNetV3 backbone architecture.}
\end{figure}

\begin{figure}[tb]
\centering
\includegraphics[width=1.0\linewidth]{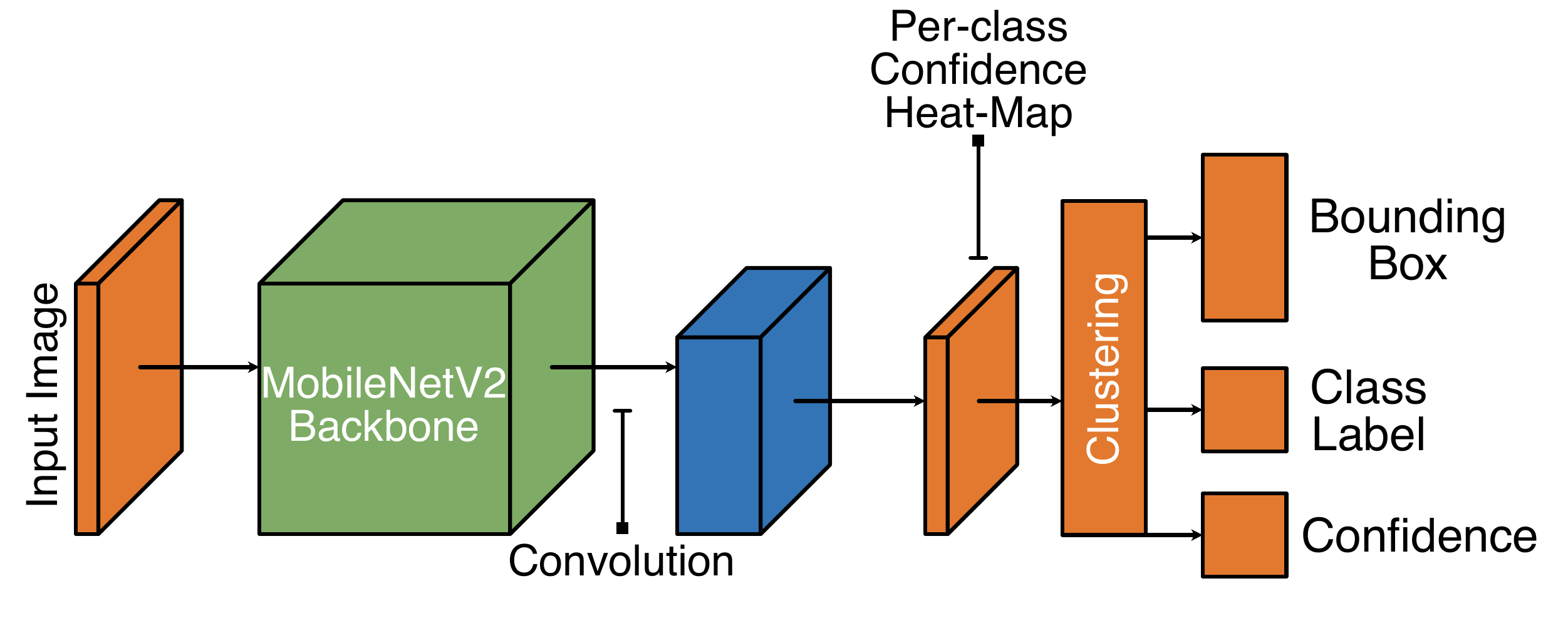}
\caption{\label{fig:ssdlite_mnet}FOMO with MobileNetV2 backbone architecture.}
\end{figure}

To address the insect detection and classification task, we use two CNNs: one devoted to the Arduino Portenta H7 and one for the GAP9 development kit. 
This choice is due to their differences, as highlighted in Section~\ref{subsec:platforms}.
For this work, we fine-tune, test, and deploy two architectures based on MobileNet~\cite{sandler2018mobilenetv2, howard2019searching} and pre-trained on the COCO dataset~\cite{cocodataset}.
Fine-tuning is a common transfer-learning~\cite{TransferLearning} technique that takes a model trained for a task, e.g., image classification, on a specific domain, e.g., the COCO challenge~\cite{cocodataset} and uses it for the same task but for a new domain exploiting the high-level knowledge obtained in the pre-trained network.

The architecture of choice for the Arduino Portenta is a FOMO MobileNet V2~\cite{sandler2018mobilenetv2} with a width multiplier of 0.35.
The network's backbone is a MobileNetV2, truncated after the 6th expands \textit{ReLU} layer, with an output feature map as big as 1/8th w.r.t to the input image. 
This feature map is then used by the \textit{faster object more object} (FOMO) detector to extract the objects of interest with their object centroids, dimensions, and classes.
The peculiarity of this detector is that it relies on a fully convolutional neural network that is trained on objects' centroids instead of bounding boxes and is designed specifically for constrained edge devices.

In the case of the FOMO MobileNetV2 network, we fine-tune four different versions that change in the input type: 160$\times$160 RGB, 160$\times$160 grayscale, 96$\times$96 RGB, and 96$\times$96 grayscale images.
The number of MACs changes depending on the input and is 16.33, 14.49, 5.88, and \SI{5.21}{\mega\nothing MAC}, respectively.
Also, the network parameters change depending on the number of channels in the input, i.e., one for grayscale images and three for RGB ones, which result in 20532 and 20820 parameters, respectively.
The fine-tuning of these networks is done using a stochastic gradient descent (SGD) optimization algorithm, running for 100 epochs with a learning rate of 0.001.

Conversely, the architecture chosen for the GAP9 is the MobileNet V3 with an SSDLite detector, which is a modification of the MobileNetV3 with SSD, suggested in~\cite{9601235} for deployment on edge devices.
In particular, the backbone of the network is a MobileNet V3~\cite{howard2019searching} architecture that has \SI{3.44}{\mega\nothing} parameters and requires \SI{584}{\mega\nothing MAC} per inference, as the architecture in~\cite{9601235}.
The detector used as the network's head is an SSDLite~\cite{sandler2018mobilenetv2} detector that is obtained, starting from the SSD model, by replacing all the regular convolutions with depth-wise separable convolutions.
Replacing the SSD layer with the SSDLite one does not affect the mAP performance, as reported in Table~\ref{tab:other_models_params}.
The SSDLite detector, which includes a non-maximal suppression layer, is up to $3.3\times$ less computationally expensive than the original SSD.
The purpose of the non-maximal suppression layer is to remove redundant and low-confidence predictions.
The network's input is a 320$\times$240 pixels image, and it produces as output several bounding boxes, together with a class label and the relative confidence, as opposed to the FOMO network, which relies on centroids.
We fine-tune the network for 300 epochs with a learning rate of 0.00025 using the SGD optimization algorithm. 
The learning rate scheduling policy considers a momentum of 0.9, with a weight decay of 0.0005, a warmup time of 10 epochs, and a minimum learning rate of 0.00005.

\subsection{Dataset}

\begin{figure}[tb]
\centering
\includegraphics[width=1.0\linewidth]{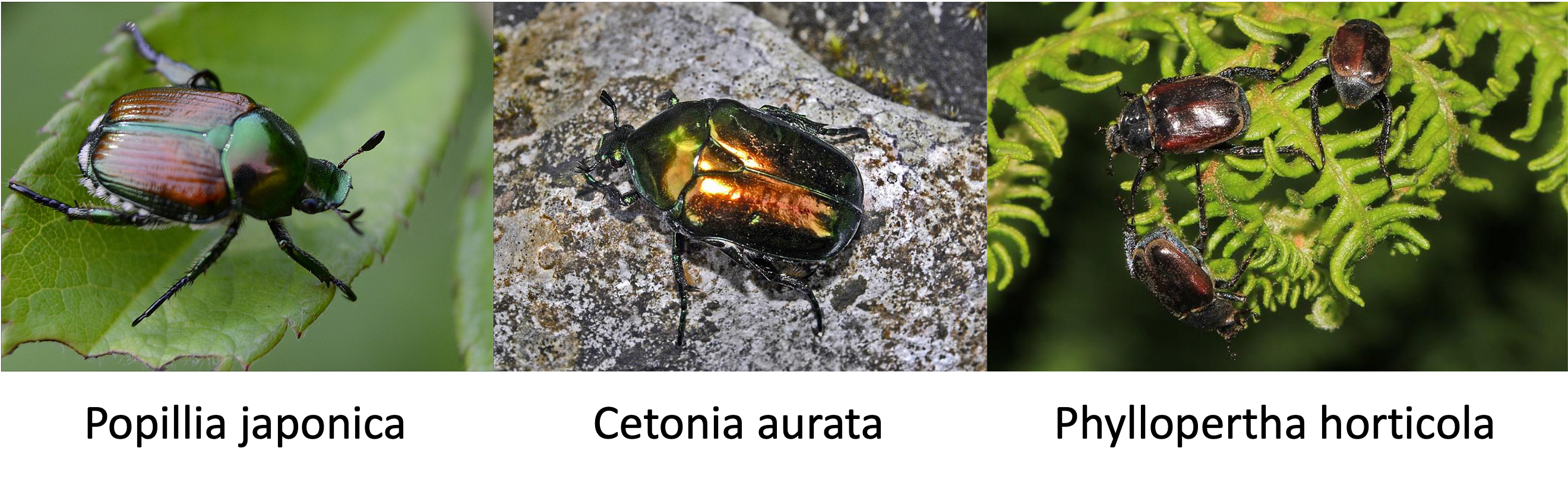}
\caption{\label{fig:datasetSamples}Samples of the three classes in our dataset~\cite{9601235}.}
\end{figure}

Our fine-tuning labeled dataset comes from~\cite{9601235} and comprises more than 3,300 images of insects, selected through a strict filtering process from an initial collection of more than 36,000 gathered through the internet. 
This filtering ensures the near-absence of duplicates that could harm the model evaluation fairness.
It contains three insect classes: Popillia japonica, Cetonia aurata, and Phyllopertha horticola. Popillia japonica is a dangerous pest insect; the other two, while similar to Popillia japonica and often mistaken for it, are not.
We have 1,422, 1,318, and 877 samples for each class, respectively.
Figure~\ref{fig:datasetSamples} reports an example of the three classes of insects.
We divided the dataset into training and testing splits, with an 80-20\% ratio.

\subsection{Deployment}

To deploy the CNNs, we leverage two tools, one for each platform, since we rely on tools that convert high-level Python code to C platform-specific code.
The first, for the Arduino Portenta board, is based on the Edge Impulse project\footnote{https://edgeimpulse.com}; the second, for the GAP9, is based on the tool developed by GWT, namely, NN-Tool.
The Edge Impulse tool provides an easy-to-use interface to deploy neural networks on many embedded devices, including the Arduino Portenta.
The tool allows the selection, fine-tuning, and deployment, in int8 or float32, of a custom or pre-designed neural network.
We deploy different variants of the networks in both fashions (int8 and float32) since the Arduino Portenta H7 SoC has hardware support for double-precision floating point operations, i.e., an FPU in each core.
Each network is exported as a \textit{tflite} model, and the inference phase is performed on the embedded device in a MicroPython-based system, thanks to OpenMV.
OpenMV, with MicroPython, provides a USB streamer and the tensor movements in the memory hierarchy with a reduced implementation effort but introduces computational overhead.
Furthermore, the OpenMV network execution does not use the RAM to store the activations during the inferences. 
The tensors need to be entirely in the L1 and L2 memory during the computation, thus avoiding the tiling from the off-chip memories but limiting the execution of the network within \SI{1}{\mega\byte} of memory.
For each network, we deploy an \textit{int8} quantized version of it, leveraging the reduced amount of memory needed for the execution of these versions of the network.

NN-Tool is the deployment tool developed by GWT to deploy neural networks on their platforms, such as GAP8 and GAP9 SoCs.
It allows the deployment starting from an \textit{onnx} or \textit{tflite} file.
We finetune the COCO pre-trained MobileNet V3 with the SSDLite detector available in Pytorch on our dataset to obtain a \textit{float32} version of the CNN.
We then convert it into an onnx file that can be used with NN-Tool.
To obtain a model that is compliant with the layers deployable with NN-Tool, such layers include convolutional, fully connected, and dropout layers or skip connections, to name a few types of operations supported. 
However, it does not support the deployment of the non-maximal suppression layer included in the architecture, and as such, we remove the operation from the onnx and apply it manually.
NN-Tool also performs the tiling and movement of tensors in the memory hierarchy to exploit all of its levels.
Since the platform has a single-precision FPU, the network is deployed in both float16 and int8 fashions.
The NE16 can be leveraged for CNN applications since the NN-tool pipeline supports deploying networks specifically tailored for this type of accelerator.
Since it can be used only with int8 networks, we deploy three architectures with NN-tool: float16, int8, and int8 with NE16 hardware acceleration.

Even if the platforms have an FPU onboard and, as such, they can perform floating point operations without relying on soft-float emulation, we perform the quantization to int8 arithmetic since it is a convenient method to reduce the memory occupation with a limited reduction in the accuracy of the network.
The quantization requires a calibration set for estimating the range of values that each tensor can assume.
Our calibration set is a subset of the training set.
\section{Results} \label{sec:results}

\subsection{Object detection performance}

\begin{figure}[t]
\centering
\includegraphics[width=1.0\linewidth]{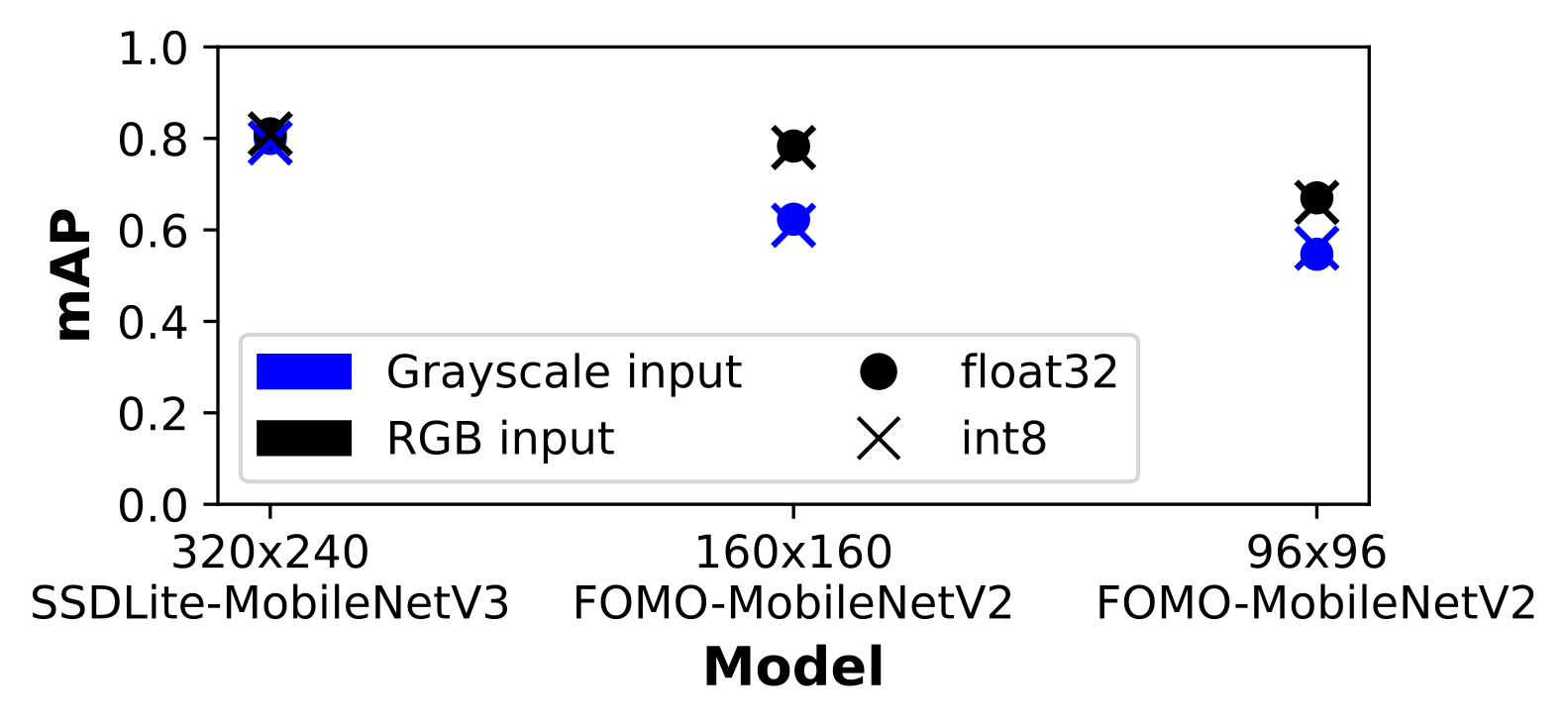}
\caption{Comparison of the mean average precision (mAP).}
\label{fig:mAP_comp}
\end{figure}

In this section, we evaluate the detection accuracy per each insect class, where the dangerous Popillia japonica is the one we want to detect.
We evaluate the network accuracy considering the mAP, a standard metric for object detection tasks, which ranges between 0 and 1.
We consider an insect correctly detected if the Intersection over Union (IoU) of the bounding box produced by the network with the ground truth is at least 0.5, thus applying the COCO standard for object detection~\cite{cocodataset}.
We test the networks on our 660 samples test set, and we explore different types of input (i.e., image color and resolution), as well as models deployed in \textit{float32}, \textit{float16}, and \textit{int8} (quantized).
Figure~\ref{fig:mAP_comp} summarizes the mAP performance of all the tested networks, with all combinations of grayscale and RGB inputs and considering \texttt{float32} and \texttt{int8} quantized versions.

Considering the models in \textit{float32} with RBG inputs, the SSDLite-MobileNetV3 is the best-performing with an mAP score of 0.80 (input size 320$\times$240), while the FOMO-MobileNetV2 achieves 0.78 in mAP (input size 160$\times$160), and finally the smallest FOMO-MobileNetV2 marks an mAP of 0.67 (input size 96$\times$96).
Then, assuming a cheaper monochrome camera, we assess the performances of the CNNs by feeding grayscale inputs but still employing \texttt{float32} arithmetic.
In this case, the SSDLite-MobileNetV3 (input size 320$\times$240), FOMO MobileNetV2 (input size 160$\times$160), and the FOMO-MobileNetV2 (input size 96$\times$96), respectively score an mAP of 0.79, 0.62, and 0.55.
The relative performances are kept the same, being the SSDLite-MobileNetV3 still the most accurate, while all three networks drop in mAP due to the grayscale input (losing from 0.01 to 0.16 in mAP).
Finally, considering the SSDLite-MobileNetV3 deployed in \texttt{float16}, to take full advantage of the single-precision FPUs available on the GAP9 SoC, we observe no drop in mAP compared with the double-precision implementation.
Then 
Instead, moving from \texttt{float32} to \texttt{int8} quantized versions of all CNNs, we see a minimal drop in performance, always lower than 3\%.

\subsection{Embedded systems performance}

In this section, we present a thorough performance assessment of all models deployed on both devices, i.e., the Arduino Portenta (STM32H74 MCU) and the GAP9 evaluation kit.
Table~\ref{tab:configurationsGAP9} reports three configurations of the GAP9 SoC, spanning the voltage from \SI{0.65}{\volt} and \SI{0.8}{\volt}, enabling different clock speeds, up to \SI{370}{\mega\hertz}.
In our experiments, we use the \textit{maximum efficiency} configuration since it achieves the longest battery duration given a target throughput.
In Table~\ref{tab:results}, we present our analysis regarding memory requirements, latency for one-image inference, and the total power consumption (including both compute unit and off-chip memories) for all CNNs/devices.
For each model, we report the data type used (\texttt{float32} and \texttt{int8} for the STM32H74, and \texttt{float16} and \texttt{int8} for the GAP9), and the input image size ($\times$3 in case of RGB images and $\times$1 for grayscale ones).
For the GAP9, the peak memory is computed considering the network's layer with the maximum requirements for input tensor, weights, and output tensor. 
Instead, for the STM32H74, the computation also includes the input image (never freed for the entire inference) and the memory footprint of a micro-python-based operating system (OS) that presses on the same \SI{1}{\mega\byte} L2 memory.
While for the GAP9 SoC, the deployment tool (NN-Tool) can organize data to exploit both on-chip L2 memory (\SI{1.6}{\mega\byte}) and the off-chip ones (up to \SI{32}{\mega\byte}), the STM32H74's tool (Edge Impulse) can not generate deployable code for CNNs requiring more than the \SI{1}{\mega\byte}-L2 on-chip memory -- this limitation is marked with $^a$ in Table~\ref{tab:results}.

\begin{table}[t]
    \small\centering
    \caption{GAP9 configurations.}
    \label{tab:configurationsGAP9}
    \resizebox{\columnwidth}{!}{%
    \begin{tabular}{llll}
    \toprule
    \textbf{Configuration}&\textbf{Voltage [\SI{}{\volt}]}&\textbf{\textbf{Freq CL [\SI{}{\mega\hertz}]}}&\textbf{\textbf{Freq FC [\SI{}{\mega\hertz}]}}\\
    \midrule
    \textbf{Min power}&0.65&150&150\\
    \textbf{Max efficiency}&0.65&240&240\\
    \textbf{Min latency}&0.80&370&370\\
    \bottomrule
    \end{tabular}
    }
\end{table}

\begin{table*}[t]
    \small
    \caption{Latency and power consumption of all networks on both STM32H74 and GAP9 MCUs.}
    \label{tab:results}
    \resizebox{\linewidth}{!}{%
    \begin{tabular}{llccccc}
    \toprule
    \textbf{Network} & \textbf{Data type} & \textbf{Input size} & \textbf{MCU} & Memory [\SI{}{\kilo\byte}] & \textbf{Latency [\SI{}{\milli\second}]} & \textbf{Total power [\SI{}{\milli\watt}]} \\
    \toprule
    FOMO-MobileNetV2$^a$ & \texttt{float32} & 96$\times$96$\times$1 & STM32H74 & 1070 & N.D. & N.D. \\
    FOMO-MobileNetV2 & \texttt{int8} & 96$\times$96$\times$1 & STM32H74 & 398 & 57 & 494 \\
    FOMO-MobileNetV2$^a$ & \texttt{float32} & 96$\times$96$\times$3 & STM32H74 & 1045 & N.D. & N.D. \\
    FOMO-MobileNetV2 & \texttt{int8} & 96$\times$96$\times$3 & STM32H74 & 416 & 62 & 498 \\
    FOMO-MobileNetV2$^a$ & \texttt{float32} & 160$\times$160$\times$1 & STM32H74 & 2654 & N.D. & N.D. \\
    FOMO-MobileNetV2 & \texttt{int8} & 160$\times$160$\times$1 & STM32H74 & 802 & 158 & 501 \\ 
    FOMO-MobileNetV2$^a$ & \texttt{float32} & 160$\times$160$\times$3 & STM32H74 & 2862 & N.D. & N.D. \\
    FOMO-MobileNetV2 & \texttt{int8} & 160$\times$160$\times$3 & STM32H74 & 854 & 169 & 499 \\
    \midrule
    SSDLite-MobileNetV3 & \texttt{float16} & 320$\times$240$\times$3 & GAP9 (CL) & 3622 & 462 & 41 \\
    SSDLite-MobileNetV3 & \texttt{int8} & 320$\times$240$\times$3 & GAP9 (CL) & 1811 & 249 & 31 \\
    SSDLite-MobileNetV3 & \texttt{int8} & 320$\times$240$\times$3 & GAP9 (NE16) & 1811 & 147 & 34 \\
    \bottomrule
    \vspace{0.01pt}
    \end{tabular}
    }
    \footnotesize{$^a$ Network not deployable (N.D.) due to its memory footprint exceeding the \SI{1}{\mega\byte} L2 available on the STM32H74. The memory footprint includes the input and output tensors and weights of the largest network layer, the input image, and the OS (in the case of the STM32H74).}
\end{table*}

On the STM32H74 MCU, we deploy the FOMO-MobilenNetV2, varying the input size from 96$\times$96$\times$1 up to 160$\times$160$\times$3.
Among the deployable CNNs on the STM32H74, the one using a monochrome 96$\times$96 image with quantized (\texttt{int8}) representation is the smallest (\SI{239}{\kilo\byte} L2) and the fastest, achieving  \SI{17.5}{frame/\second}.
All the deployable networks require less than \SI{501}{\milli\watt}, primarily due to MCU's power consumption, which, in typical operating conditions, consumes $\sim$\SI{480}{\milli\watt} (using only the M7 core).

On the GAP9 SoC, we deploy the SSDLite-MobileNetV3 with a fixed input size of 320$\times$240$\times$3, considering three execution conditions: \textit{i}) \texttt{float16} running on general-purpose CL (with FPU hardware support), \textit{ii}) \texttt{int8} quantized running on the CL, and \textit{iii}) with \texttt{int8} but exploiting the NE16 convolutional accelerator.
As expected, the \texttt{float16} version is the most memory demanding, requiring up to $\sim$\SI{7}{\mega\byte} of memory, and the slowest. 
Despite the platform's FPU hardware support, the \textit{float16} network reaches only \SI{2.1}{frame/\second}, consuming \SI{40.6}{\milli\watt}.
Instead, employing the NE16 accelerator with the \texttt{int8} data type, requires up to \SI{3.4}{\mega\byte} of memory and reaches up to \SI{6.8}{frame/\second} within only \SI{34}{\milli\watt}.

\begin{figure}[tb]
\centering
\includegraphics[width=1.\linewidth]{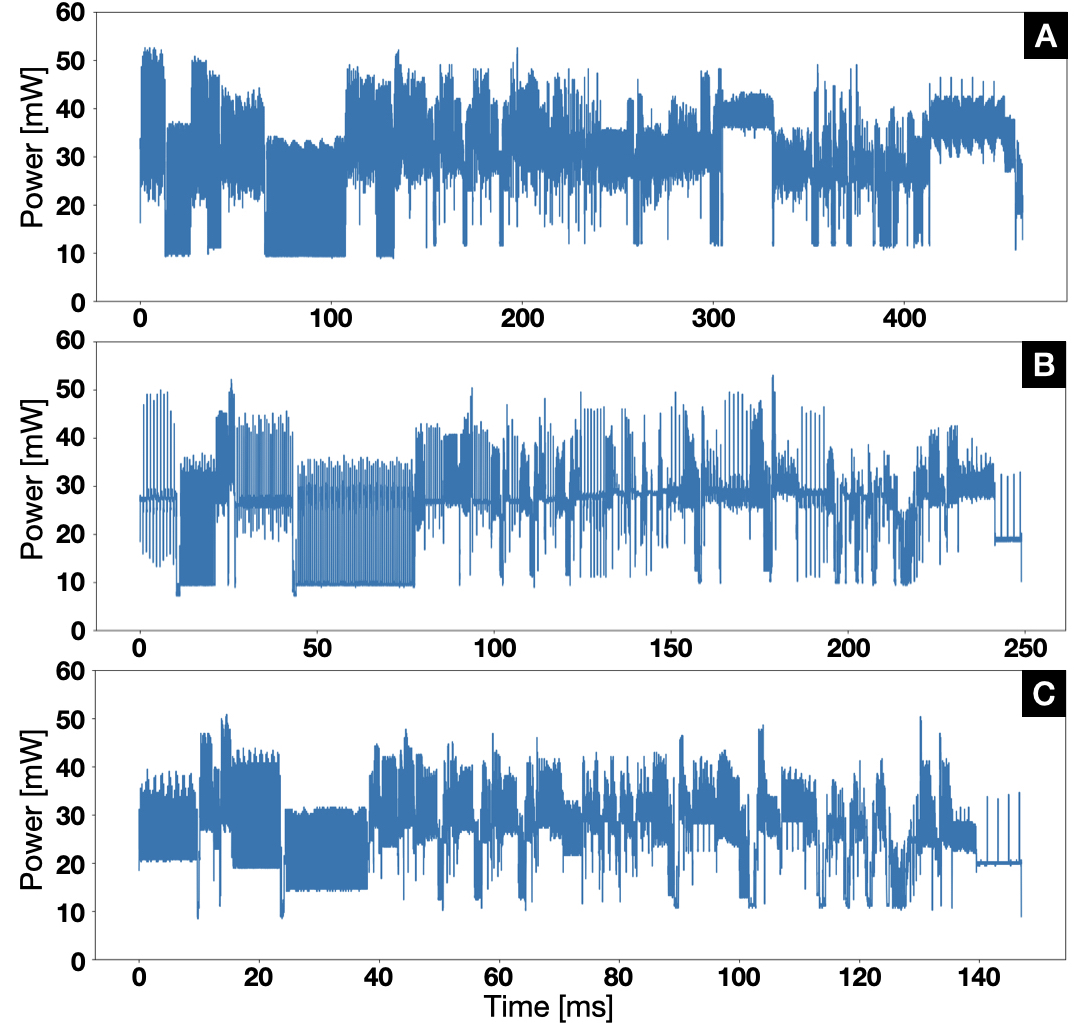}
\caption{Power waveforms for the three deployed networks on the GAP9, i.e., A) \textit{float16}, B) \textit{int8} running on the CL, and C) \textit{int8} running on the NE16 accelerator.}
\label{fig:powerGAP9}
\end{figure}

Finally, in Figure~\ref{fig:powerGAP9}, we report the waveform of the power consumption of the GAP9 SoC in three cases, i.e., with the \textit{float16} version, the \textit{int8} executed on the general-purpose multicore cluster, and the \textit{int8} running on the NE16 accelerator, respectively in Figure~\ref{fig:powerGAP9}-A, -B and -C.
All three waveforms show small periods of idleness (up to \SI{41}{\milli\second} in total), in particular, at the beginning of the execution of the network, where the tensor movements between memories do not overlap with the execution of the network. 
The idle periods are reduced going forward in the execution since the size of the tensors reduces.

\subsection{Discussion}

Compared to the best-in-class network, the RetinaNet-ResNet101-FPN full-precision~\cite{9601235}, our best model drops only 15\% in accuracy evaluating it with the mAP but reduces the number of operations by 299$\times$ and the memory required by 14.9$\times$ allowing the deployment of the system on an ultra-low power embedded system such as our GAP9-based board.
On the other hand, the FOMO MobileNetV2 reduces the number of operations by $\sim$20000$\times$ and the memory by $\sim$2500$\times$ w.r.t. the RetinaNet-ResNet101-FPN full-precision~\cite{9601235} with a reduction of mAP of 0.27.
The impressive parameters' reduction of both our networks, the FOMO MobileNet V2 and the SSDLite-MobileNetV3, allows the deployment of our system in traps as an IoT device to detect insects such as Popillia japonica, Cetonia aurata, and Phyllopertha horticola.
We envision a battery-powered insect detection system that uses one of our platforms, the Arduino Portenta H7 with the FOMO MobileNetV2 or the GAP9-based development kit running our MobileNetV3 with SSDLite, an ultra-low-power camera such as the HIMAX HM-01B0 for image acquisition, and a LoRaWAN module, e.g, the HTCC-AB01 board by CubeCell, for the transmission of the detections.
We estimate the duration of the battery considering a \SI{1000}{\milli\ampere\hour} battery rated at \SI{3.7}{\volt}, which can provide up to \SI{7.4}{\watt\hour}, equivalent to \SI{13.3}{\kilo\joule}, to our system.

\begin{figure}[tb]
\centering
\includegraphics[width=1.\linewidth]{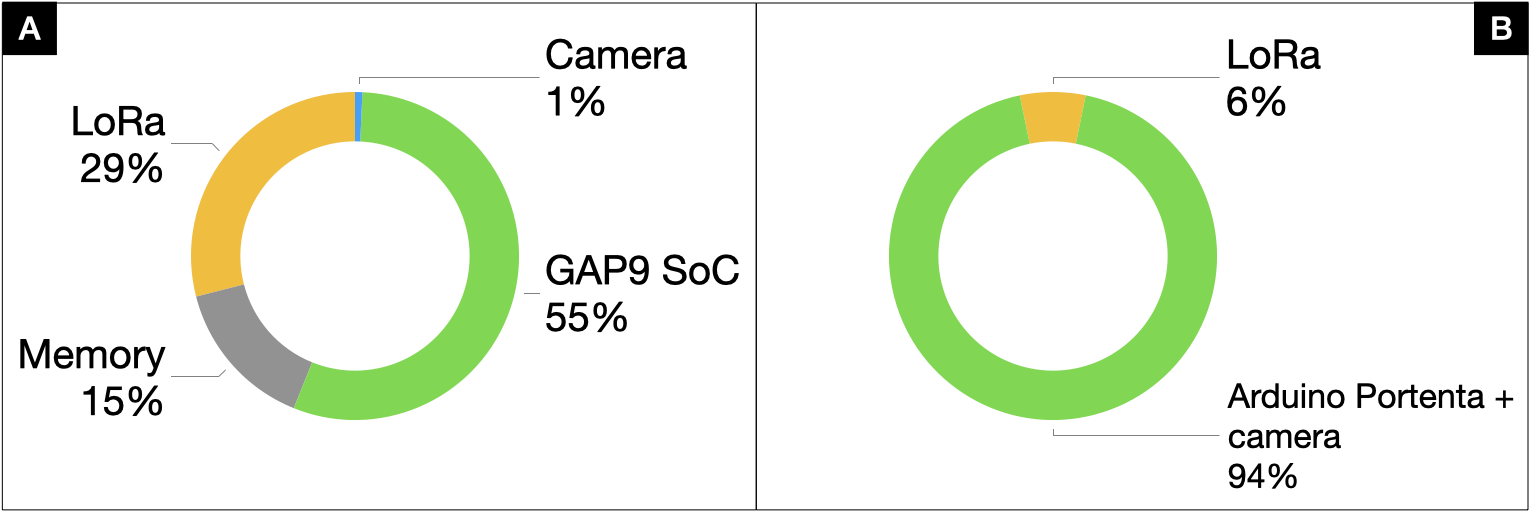}
\caption{Power breakdown considering the image acquisition, the processing, and the transmission. A) GAP9 board, B) Arduino Portenta.}
\label{fig:powerBrakedown}
\end{figure}

On the one hand, the energy consumption can be subdivided into \SI{3.82}{\milli\joule} for the GAP9 SoC, \SI{1.03}{\milli\joule} for the memory, \SI{0.05}{\milli\joule} for the camera, and \SI{2}{\milli\joule} for the LoRaWAN transmission system, considering two bytes sent each time.
We highlight that the LoRaWAN transmission module is not strictly required in the case of traps since we can also rely on a visual signaling system, such as an LED positioned near the trap, that signals the detection of the Popillia japonica.
As a consequence, our GAP9-based system consumes \SI{6.9}{\milli\joule} with the LoRa module or \SI{4.9}{\milli\joule} without the LoRa module.
The system for the Portenta board consumes \SI{31}{\milli\joule} with the LoRa module or \SI{29}{\milli\joule} without.
Furthermore, we need to consider the energy consumption in deep sleep mode for the GAP9 SoC, that is \SI{40}{\joule} per day, and for the Arduino Portenta, which is \SI{214}{\joule} per day.
As such, the battery can last up to \SI{267}{\nothing} days with the LoRa transmission module or more than \SI{283}{\nothing} days in the case of LED-based signaling in the case of the GAP9 board.
For the Arduino Portenta, it reduces to \SI{51}{\nothing} days with the LoRa transmissions and \SI{58}{\nothing} without it.
Figure~\ref{fig:powerBrakedown} reports the power breakdown of the system described above.
With the system described above and running it once every \SI{1}{\minute}, we expect up to 267 working days on the GAP9 SoC without recharging nor changing the battery or up to 58 days on the Arduino Portenta.

Either our two detection systems can be deployed aboard nano-drones since they can fit the power envelope ($\sim$\SI{10}{\watt} including also the motors) and the $\sim$\SI{10}{\gram} payload available onboard. 
This allows the envisioning of a pest-detection system based on a nano-drone that flies into greenhouses and provides localized information for the use of pesticides, drastically reducing their use related to widespread adoption.
\section{Conclusions} \label{sec:conclusion}

This work presents a novel hardware-software design for image-based pest detection, deployable in battery-powered smart traps and aboard ultra-constrained nano-drones. 
We present two system designs featuring two ultra-low power embedded devices, i.e., a widely used dual-core Arduino Portenta H7 and a novel multi-core GWT GAP9 SoC featuring the NE16 hardware accelerator.
Given these two devices' significant memory and computing power differences, we explore two alternative SoA CNNs.
On the Portenta, we deploy a lightweight FOMO-MobileNetV2 (\SI{5.88}{\mega MAC/inference}), capable of reaching 0.66 mAP on detecting the Popilla japonica bug, running at \SI{16.1}{frame/\second} and consuming \SI{498}{\milli\watt}.
While on the GAP9 SoC, we deploy a more complex SSDLite-MobileNetV3 CNN (\SI{584}{\mega MAC/inference}), scoring an mAP of 0.79 with a throughput of \SI{6.8}{frame/\second} at \SI{33}{\milli\watt}.
With our hardware-software codesign, we present a fine-tuning procedure with a custom dataset for the detection task, an 8-bit quantization stage for efficient exploitation of the two SoCs, and finally, the implementation and deployment of our workloads.
Compared to the huge first-in-class RetinaNet-ResNet101-FPN (\SI{174850}{\mega MAC/inference}), our best model drops only 15\% in mAP, paving the way toward autonomous palm-sized drones capable of lightweight and precise pest detection.

\bibliographystyle{./IEEEtran}
\bibliography{biblio}

\end{document}